\documentclass[sigconf]{acmart}
\setcopyright{none}
\renewcommand\footnotetextcopyrightpermission[1]{} % removes footnote with conference information in first column
\AtBeginDocument{%
  }

\usepackage{bbding}
\usepackage{float}
\usepackage{natbib}
\usepackage{multirow}
\usepackage{subcaption}
\usepackage{amsmath}
\usepackage{empheq}
\usepackage[noend]{algpseudocode}
\usepackage{algorithmicx,algorithm}
\usepackage{makecell}
\usepackage{multirow}
\usepackage{appendix}
\usepackage{booktabs}
\usepackage[capitalize]{cleveref}
\crefname{section}{Sec.}{Secs.}
\Crefname{section}{Section}{Sections}
\crefname{figure}{Fig.}{Figs.}
\Crefname{figure}{Figure}{Figures}
\Crefname{table}{Table}{Tables}
\crefname{table}{Tab.}{Tabs.}
\Crefname{algorithm}{Algorithm.}{Algorithms.}
\crefname{algorithm}{Appendix.}{Appendixes.}
\crefname{appendix}{App.}{Apps.}

\usepackage{booktabs}
\usepackage{tabularx}
\usepackage{enumitem}

\setlist{nosep}

\usepackage{multirow}
\usepackage[edges]{forest}
\definecolor{customgreen}{HTML}{70AD47} % Define the color with a custom name
\definecolor{customorange}{HTML}{E6990E} % Define the color with a custom name
\usepackage{url}
\definecolor{myblue}{HTML}{70AD47}
\definecolor{lightblue}{HTML}{fcecd0}

\usepackage{tikz}
\usetikzlibrary{arrows.meta, positioning}
\definecolor{SingleColor}{HTML}{FFF2CC}
\definecolor{MultiColor}{HTML}{D9EAD3}
\definecolor{ArrowBlue}{HTML}{B4D7FE}

\begin{document}

\title{Towards Agentic Recommender Systems in the Era of Multimodal Large Language Models}

\author{
 \textbf{Chengkai Huang}\textsuperscript{1}, 
 \textbf{Junda Wu}\textsuperscript{2}, 
 \textbf{Yu Xia}\textsuperscript{2}, 
 \textbf{Zixu Yu}\textsuperscript{2}, 
 \textbf{Ruhan Wang}\textsuperscript{3}, 
 \textbf{Tong Yu}\textsuperscript{4}, 
 \textbf{Ruiyi Zhang}\textsuperscript{4}, 
 \textbf{Ryan A. Rossi}\textsuperscript{4}, 
 \textbf{Branislav Kveton}\textsuperscript{4}, 
 \textbf{Dongruo Zhou}\textsuperscript{3}, 
 \textbf{Julian McAuley}\textsuperscript{2}, 
 \textbf{Lina Yao}\textsuperscript{1,5}
\\[6pt] 
 \textsuperscript{1}University of New South Wales,
 \textsuperscript{2}University of California San Diego,
\\
 \textsuperscript{3}Indiana University,
 \textsuperscript{4}Adobe Research, 
 \textsuperscript{5}CSIRO’s Data61
\\
\small{\texttt{\textbraceleft chengkai.huang1, lina.yao\textbraceright @unsw.edu.au, \textbraceleft juw069, yux078, ziy040, jmcauley\textbraceright @ucsd.edu, \textbraceleft ruhwang, dz13\textbraceright @iu.edu, \textbraceleft tyu, ruizhang, rrossi, kveton\textbraceright @adobe.com}}
}

\renewcommand{\shortauthors}{Huang et al.}

\begin{abstract}
Recent breakthroughs in Large Language Models (LLMs) have led to the emergence of agentic AI systems that extend beyond the capabilities of standalone models. By empowering LLMs to perceive external environments, integrate multimodal information, and interact with various tools, these agentic systems exhibit greater autonomy and adaptability across complex tasks. This evolution brings new opportunities to recommender systems (RS): LLM-based Agentic RS (LLM-ARS) can offer more interactive, context-aware, and proactive recommendations, potentially reshaping the user experience and broadening the application scope of RS.
Despite promising early results, fundamental challenges remain, including how to effectively incorporate external knowledge, balance autonomy with controllability, and evaluate performance in dynamic, multimodal settings. In this perspective paper, we first present a systematic analysis of LLM-ARS: (1) clarifying core concepts and architectures; (2) highlighting how agentic capabilities—such as planning, memory, and multimodal reasoning—can enhance recommendation quality; and (3) outlining key research questions in areas such as safety, efficiency, and lifelong personalization. We also discuss open problems and future directions, arguing that LLM-ARS will drive the next wave of RS innovation. Ultimately, we foresee a paradigm shift toward intelligent, autonomous, and collaborative recommendation experiences that more closely align with users’ evolving needs and complex decision-making processes.
\end{abstract}



\maketitle

\section{Introduction}\label{sec:intro}

% Rec background
With the rapid growth of online services, recommender systems (RS) have become essential for addressing users' information needs and alleviating information overload \cite{RicciRS15,ZhangYST19}. 
These systems provide personalized recommendations across various domains, including e-commerce, movies, music, etc. 
Despite the diversity of recommendation tasks such as top-K recommendation and sequential recommendation, 
the core objective remains consistent: to predict a user’s preferences for each candidate item and generate a ranked list tailored to the user \cite{survey1}.

% shortcomings: 
% open-domain knowledge, implicit modeling (personalization/alignment), lack interactive features
% However, traditional recommender systems face several significant limitations in meeting diverse user needs. First, these systems are often domain-specific, built on discrete ID features limited to particular contexts, and lacking the open-domain knowledge needed for cross-domain applicability. This constraint hinders the system's ability to fully understand user interests and item content across various domains and platforms, resulting in suboptimal recommendation performance. 
% Additionally, traditional systems typically optimize for specific user behaviors, such as clicks or purchases, based on historical data, where user preferences and underlying motivations are modeled implicitly. This approach can reduce recommendation transparency and may fail to capture the complex, dynamic intentions of users in different contexts. 
% Furthermore, these systems usually lack interactive features, meaning users cannot actively guide recommendations by providing direct, natural language input, which limits the degree of personalization.

% limitations
However, current RSs still face several significant limitations in meeting diverse user needs.
First, current RSs typically rely on ID-based features that work only within specific domains or platforms. Their inability to integrate open-domain knowledge, such as common sense reasoning and cross-platform behavioral patterns, significantly constrains their capacity to interpret and model user interests in a broader context.
Second, current methods typically optimize well-defined engagement metrics derived from historical interaction data (e.g., click-through rates and purchase histories). Although such methods can be effective for localized objective functions, they often conflate observable behaviors with latent user intent, since implicit feedback mechanisms cannot distinguish transient actions from enduring preferences. Consequently, these models exhibit two major limitations: (i) lack of transparency regarding preference attribution, which impairs interpretability, and (ii) oversimplification of the multifaceted motivations that guide user behavior, especially in scenarios requiring temporal or situational adaptation. As a result, these implicit modeling frameworks fail to capture the causal relationships between dynamic user states and subsequent decision-making processes.
Finally, most traditional RSs operate in a largely static, one-directional manner, providing users with minimal opportunities to iteratively refine suggestions through natural language or real-time feedback. This unidirectional flow diverges from established human-computer interaction principles, which emphasize interactive, adaptive dialogue to uncover user preferences. Although conversational RSs have begun to address this issue, they remain limited in their ability to integrate open-ended natural language understanding with personalized ranking, particularly in scenarios that require multiple rounds of clarification to resolve ambiguous user queries.
% CRS interface
% Conventional recommender systems lack interactive customization options. By leveraging LLMs, users can provide explicit, natural language instructions to adjust and personalize recommendations on the spot, leading to an experience that is both more transparent and adaptable to specific user requirements.

% Agent bg
% Recent progress in Large Language Models (LLMs) and Multimodal Large Language Models (MLLMs) has significantly enhanced their capabilities in language comprehension and cognitive processing to new heights \cite{GPT4report,jaech2024openai,guo2025deepseek}.
% % Agent intro
% With improved natural language comprehension and enhanced reasoning capabilities, (M)LLM-based agents can effectively interpret and utilize human language, develop detailed plans, and execute complex tasks. These breakthroughs provide new opportunities for researchers to tackle longstanding challenges within RSs, enhancing their adaptability, personalization, and user-centric focus.
Recent advances in Large Language Models (LLMs) and Multimodal LLMs (MLLMs) have greatly improved language comprehension and cognitive processing \cite{GPT4report,jaech2024openai}. With stronger reasoning and planning abilities, (M)LLM-based agents can interpret human language, devise strategies, and execute complex tasks. These breakthroughs offer new avenues for enhancing RSs' adaptability, personalization, and user-centricity.
% As a new generation of intelligent techniques, LLMs present unprecedented opportunities to transform RS research and applications. 
The rapid surge in LLM-driven RS research is evident from the 290 references cited in a recent survey on this topic \cite{survey1,survey2,huang2024foundation}, along with numerous influential papers in the field (e.g., \cite{Shashank2025}). The existing work on applying LLMs to RS, however, has mostly focused on applying LLMs to improve the current RSs. 
Furthermore, \textbf{the existing works have underexplored the important question of how LLMs or LLM agents would impact the future of RS in the long run.} 
We argue that LLM-based Agentic Recommender Systems (LLM-ARS) present a promising research direction, offering new perspectives on autonomy, adaptability, and interactive decision-making in recommendation.
To unlock the full potential of LLM-ARS, it is crucial to address several open questions, including how to harness agentic capabilities (e.g., planning, collaboration, roleplaying) to improve user modeling and system decision-making, and how to balance autonomy with controllability to ensure safe, transparent interactions. We offer a more detailed discussion of these challenges and key research questions in Section \ref{sec:RQs}, where we highlight the most pressing issues and outline possible solutions.
% In this paper, we propose the following \textbf{key questions:} Why are LLM-ARS a key direction for future development? What are the key components of LLM-ARS under our expectation? Building on these questions,
% we present a comprehensive survey of LLM-ARS.
% We structure our survey from two perspectives:
% From the MLLM agent perspective, we examine the design and key components of LLM-based agents in recommendation systems, focusing on their reasoning, memory, planning, and adaptability. From the Recommender Systems perspective, we analyze how LLM-based agents can enhance existing recommendation tasks and improve overall system performance.

% We conduct the first perspective paper in agentic RS with (M)LLMs. We first give some preliminaries and backgrounds on this direction (§2). Then, we claim why LLM-ARSs are important. (§3) and formulate LLM-ARS in (§4). Then we analyze the LLM-ARS from the agent's perspective (§5). Then we propose the key research questions from RS perspectives. Aiming to these questions, we further extend in-depth comparisons with a discussion for insights (§7 and §8). For unsolved and need-to-addressed questions, we highlight them in the open problem and opportunities section (§9).

\textbf{We present the first perspective paper on ARS powered by (M)LLMs}. We begin with preliminaries and background on this emerging direction (§2), followed by a discussion on the significance of LLM-ARS (§3) and a formal problem formulation (§4). Next, we analyze LLM-ARS from an agentic perspective (§5) and introduce key research questions from the RS standpoint (§6). To address these questions, we provide in-depth comparisons and discussions, offering insights into the field (§7 and §8). Finally, we highlight open problems and future opportunities that require further exploration (§9). In summary, our key contributions in this perspective paper are as follows:

\begin{itemize}[left=0pt]
    \item  We position LLM-ARS within the broader trajectory of RS development by introducing a four-level evolution, emphasizing the shift from static, one-way recommendation toward agentic paradigms that support autonomy and interactive decision-making.
    \item We propose a formal task formulation for LLM-ARS, detailing the core components—user profiling, planning, memory, and action—that together enable continuous adaptation and proactive recommendations.
    \item We identify critical research questions and open problems of how to harness agentic capabilities (e.g., planning, roleplaying, collaboration) to improve user modeling, system decision-making, and overall recommendation effectiveness.
\end{itemize}

% timeline graph
\section{Preliminary and Background}

\begin{figure*}
\begin{tikzpicture}[
  font=\small,
  boxSingle/.style={
    draw,
    thick,
    fill=SingleColor,
    rounded corners=2pt,
    inner sep=3pt,
    align=center,
    text width=1.7cm,
    anchor=west
  },
  boxMulti/.style={
    draw,
    thick,
    fill=MultiColor,
    rounded corners=2pt,
    inner sep=3pt,
    align=center,
    text width=1.7cm,
    anchor=west
  },
  scale=1.0,
  every node/.style={transform shape}
]

\shade[top color=white, bottom color=ArrowBlue]
      (-2.4,1.0) rectangle (-1.6,-8.3);

\draw[-{Latex[length=3mm,width=2mm]}, very thick, color=blue!70]
      (-2.0,2.0) -- (-2.0,-8.3);

\node[
  draw,thick,fill=SingleColor,rounded corners=2pt,
  anchor=south, minimum width=3cm,align=center
] at (1.5,2.1) {\textbf{Single Agent}};

\node[
  draw,thick,fill=MultiColor,rounded corners=2pt,
  anchor=south, minimum width=3cm,align=center
] at (7.8,2.1) {\textbf{Multi Agent}};

\draw[dashed, thick] (5.2,2.5) -- (5.2,-8.0);

\node[anchor=west] at (-1.5, 1.5) {\textbf{2021}};
\draw[dashed] (-1.0, 1.2) -- (10, 1.2);

\node[anchor=west] at (-1.5, -0.2) {\textbf{2022}};
\draw[dashed] (-1.0, -0.5) -- (10, -0.5);

\node[anchor=west] at (-1.5, -2.1) {\textbf{2023}};
\draw[dashed] (-1.0, -2.4) -- (10, -2.4);

\node[anchor=west] at (-1.5, -3.9) {\textbf{2024}};
\draw[dashed] (-1.0, -4.2) -- (10, -4.2);

\def\xAone{0.5}
\def\xAtwo{2.8}
\def\xMone{5.6}
\def\xMtwo{7.9}

% == 2021 ==

% Single Agent
\node[boxSingle] at (\xAone, 0.6) {WebGPT};
\node[boxSingle] at (\xAtwo, 0.6) {Pangu};

% == 2022 ==

% Single Agent
\node[boxSingle] at (\xAone, -1.1) {WebShop};
\node[boxSingle] at (\xAtwo, -1.1) {Reflexion};
\node[boxSingle] at (\xAone, -1.9) {ReAct};
\node[boxSingle] at (\xAtwo, -1.9) {RAP};

% (2022 MultiAgent?)

% == 2023 ==

% Single
\node[boxSingle] at (\xAone, -2.8) {AutoGPT};
\node[boxSingle] at (\xAtwo, -2.8) {Retroformer};
\node[boxSingle] at (\xAone, -3.6) {SayCanPay};
\node[boxSingle] at (\xAtwo, -3.6) {LATS};
\node[boxSingle] at (\xAone, -4.6) {AVIS};
\node[boxSingle] at (\xAtwo, -4.6) {ToolChain};

\node[boxMulti] at (\xMone, -2.8) {Camel};
\node[boxMulti] at (\xMtwo, -2.8) {Agents};
\node[boxMulti] at (\xMone, -3.6) {AutoGen};
\node[boxMulti] at (\xMtwo, -3.6) {MetaGPT};
\node[boxMulti] at (\xMone, -4.6) {OpenAgents};
\node[boxMulti] at (\xMtwo, -4.6) {AgentVerse};

\node[boxSingle] at (\xAone, -5.5) {WebArena};
\node[boxSingle] at (\xAtwo, -5.5) {RecMind};
\node[boxSingle] at (\xAone, -6.3) {CLOVA};
\node[boxSingle] at (\xAtwo, -6.3) {Agent-FLAN};
\node[boxSingle] at (\xAone, -7.1) {AgentKit};
\node[boxSingle] at (\xAtwo, -7.1) {AIOS};
\node[boxSingle] at (\xAone, -7.9) {KnowAgent};
\node[boxSingle] at (\xAtwo, -7.9) {AvaTaR};

\node[boxMulti] at (\xMone, -5.5) {AgentLite};
\node[boxMulti] at (\xMtwo, -5.5) {ChatEval};
\node[boxMulti] at (\xMone, -6.3) {CoMM};
\node[boxMulti] at (\xMtwo, -6.3) {AgentDebate};
\node[boxMulti] at (\xMone, -7.1) {ChatDev};
\node[boxMulti] at (\xMtwo, -7.1) {MACRec};
\node[boxMulti] at (\xMone, -7.9) {MAGIS};
\node[boxMulti] at (\xMtwo, -7.9) {MASTER};

\end{tikzpicture}
\caption{The rising trend in the research field of LLM-based Agents. We categorize current work into single-agent and multi-agent categories.}
\label{fig:enter-label}
\end{figure*}
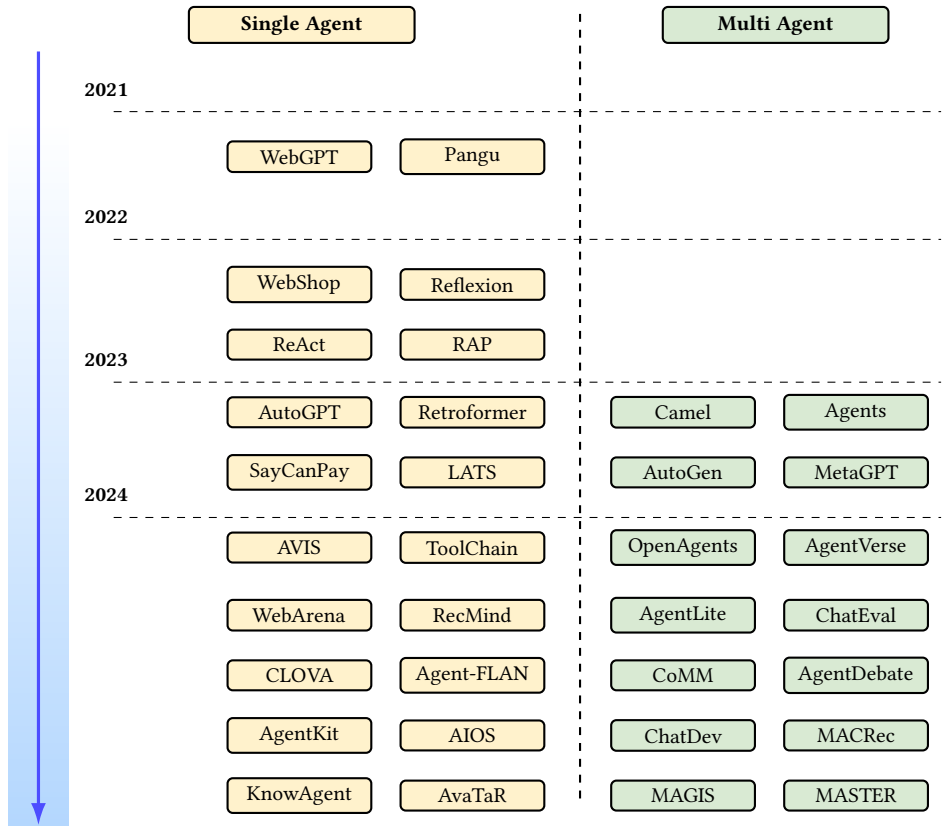

The rapid evolution of LLM-based AI has spurred significant advancements in Agent AI, fundamentally reshaping how systems interact with complex environments. In recent years, researchers have equipped LLM agents with core components—memory, planning, reasoning, tool utilization, and action execution—that are essential for autonomous decision-making and dynamic interaction \cite{durante2024agent}. 
The following subsections together with Figure \ref{fig:enter-label} provide an overview of the recent developments in both single-agent and multi-agent frameworks.
%Large language model (LLM) based AI systems have become pivotal in recent advancements of Agent AI, enabling sophisticated, dynamic applications across diverse domains. These agents are built upon core components—memory, planning, reasoning, tool utilization, and action—that collectively satisfy the essential requirements for agents to understand, reason about, and interact with their environments effectively \cite{durante2024agent}.

\subsection{LLM-based Single-Agent Systems}

Single-agent systems leverage a unified model that integrates multiple interdependent modules.\footnote{\url{https://github.com/huggingface/smolagents}}\footnote{\url{https://www.langchain.com/langgraph}}
The memory component acts as a structured repository that stores and retrieves contextually relevant information, such as user preferences and historical interactions \cite{zhang2024survey}. This persistent memory is crucial for maintaining coherent, long-term interactions and forms the foundation for personalization in recommendation settings.
The planning module is closely linked with advanced reasoning capabilities. Recent research has identified approaches such as task decomposition, multi-plan selection, external module-aided planning, reflection and refinement, and memory-augmented planning \cite{huang2024understanding}.
These techniques enable an agent to break down complex tasks, select and refine strategies based on evolving contexts, and leverage external knowledge sources.
Integrated reasoning further enhances decision-making by allowing the system to adapt dynamically to novel scenarios. Frameworks like ReAct \cite{yao2022react} and Reflexion \cite{shinn2023reflexion} exemplify how interleaving reasoning with concrete actions—such as web-browsing or tool invocation—can significantly improve system robustness and adaptability.
Beyond internal cognitive processes, these agents increasingly rely on tool utilization to interface with external data and services. Systems like WebGPT \cite{nakano2021webgpt} illustrate the effectiveness of using external modules (e.g., web search engines) to retrieve real-time information. 
Other works, such as Retroformer \cite{yao2023retroformer} and AvaTaR \cite{wu2024avatar}, further optimize these interactions through policy gradient optimization and contrastive reasoning, respectively, to fine-tune tool usage and enhance performance over time.

\subsection{LLM-based Multi-Agent Systems}

In contrast, LLM-based multi-agent systems emphasize collaboration among diverse autonomous agents. These systems are designed to mimic complex human workflows by facilitating inter-agent communication, task specialization, and coordinated decision-making.
Frameworks such as CAMEL \cite{li2023camel} and AutoGen \cite{wu2023autogen} demonstrate how agents with distinct roles can interact to solve problems more efficiently than a single, monolithic agent. By assigning specialized functions—ranging from ideation and planning to evaluation—these frameworks enable a division of labor that enhances overall system capability and flexibility.
Further advancements are seen in approaches like MetaGPT \cite{hong2023metagpt} and AgentLite \cite{liu2024agentlite}, which incorporate meta-programming techniques and lightweight libraries to dynamically allocate roles and coordinate complex workflows. These structured interactions not only improve task efficiency but also offer robustness in dynamic problem-solving environments.
Recent developments also include systems such as ChatEval \cite{chan2023chateval} and ChatDev \cite{qian2023communicative}, which leverage inter-agent debate and evaluative feedback to produce more nuanced and reliable outputs. This human-like discussion among agents is particularly beneficial in open-ended natural language generation tasks and complex software development processes.

% \begin{figure*}[ht]
%     \centering
%     \includegraphics[width=1\linewidth,height=1\textheight,keepaspectratio]{figures/agent_timeline.pdf}
%     \caption{The rising trend in the research field of LLM-based Agents. We categorize current work into single-agent and multi-agent categories.}
%     \label{fig:enter-label}
%     % \vspace{-2em}
% \end{figure*}
% \input{tables/evolution_table}

\section{Why Agentic Recommender Systems Now?}

% LLMs enhance but Won’t Replace Recommender systems?

% 1. How to utilize LLMs to rec?
% 1) In-context learning
% 2) Tuning
% -> Agent

% 2. Information modalities: From pure ID and text modality to multiple modalities

% 3. User interface: more personalized and conversational rec, from passive to proactive

Recent advances in RSs have largely focused on enhancing interaction capabilities, 
with most research efforts still operating at the Advanced RSs (Level 1) and Intelligent RSs (Level 2) stages as shown in Table \ref{tab:evolution}. 
However, they remain fundamentally reactive, relying on predefined model architectures and user-driven feedback loops.
The next frontier, Agentic RSs (Level 3), aims to move beyond reactive engagement to autonomous, adaptive, and proactive recommendation strategies,
which is increasingly feasible due to recent breakthroughs in (M)LLMs. We identify three key factors:
\begin{table*}[ht]
\small
    \centering
    \renewcommand{\arraystretch}{1.3} % 增加行距
    % \begin{tabular}{|l|l|p{4cm}|p{7cm}|}
    \begin{tabular}{l|p{2cm}|p{3cm}|p{11cm}}
        \toprule
        \textbf{Level} & \textbf{Name} & \textbf{Description} & \textbf{Key Characteristics} \\
        \hline
        \textit{0} & Traditional Recommender Systems & 
        Systems rely on static algorithms and historical data to suggest items. & 
             $\bullet$ \textbf{Rule-Based Processing:} Uses fixed rules, collaborative filtering, or content-based methods.
             $\bullet$ \textbf{Limited Contextual Understanding:} Operates solely on past user behavior without real-time adjustments.
             $\bullet$ \textbf{One-Way Interaction:} Provides recommendations in a non-interactive, one-off manner.
 \\
        \hline
        \textit{1} & Advanced Recommender Systems & 
       Deep learning advances enhance personalization with historical and real-time data.
        % Advances in deep learning enable leveraging both historical and real-time data for improved personalization.
        & 
             $\bullet$ \textbf{Data-Driven Adaptation:} Uses learning models to update recommendations based on new information.
             $\bullet$ \textbf{Feedback Integration:} Incorporates user feedback to refine suggestions over time.
             $\bullet$ \textbf{Enhanced Personalization:} Provides more accurate and context-aware recommendations while following predefined model structures.
  \\
        \hline
        \textit{2} & Intelligent Recommender Systems & 
        These systems actively engage users to refine their understanding of preferences. & 

             $\bullet$ \textbf{Interactive Engagement:} Initiates clarifying dialogues and solicits additional input.
             $\bullet$ \textbf{Multi-Modal Input Processing:} Integrates inputs beyond text (e.g., images, behavioral signals).
             $\bullet$ \textbf{Dynamic Adaptation:} Adjusts recommendations in real-time based on user context.
   \\
        \hline
       \textit{3} & Agentic Recommender Systems & 
        Fully autonomous agents that not only provide recommendations but also self-improve and evolve. & 

             $\bullet$ \textbf{Autonomous Decision-Making:} Uses planning and optimization to proactively shape recommendation strategies.
             $\bullet$ \textbf{Continuous Self-Evolution:} Updates models and behaviors based on internal and external feedback.
             $\bullet$ \textbf{Comprehensive Memory \& Multi-Modal Perception:} Integrates long-term user data, contextual cues, and multiple input types.
             $\bullet$ \textbf{Proactive and Reactive Interactions:} Balances immediate responses with strategic actions.
             \\
        \hline
    \end{tabular}
    \caption{Four-Level Evolution of Recommender Systems: In this study, we categorize RSs into four levels based on their adaptability and interaction capabilities. Traditional RSs rely on static algorithms and historical data, while advanced RSs leverage deep learning for real-time personalization. Intelligent RSs engage users interactively, and agentic RSs autonomously evolve and optimize recommendations.}
    \label{tab:evolution}
    \vspace{-1em}
\end{table*}

\begin{itemize}[left=0pt]
    \item \textbf{Leveraging (M)LLMs for Recommendation:} The integration of LLMs introduces agent-like capabilities such as planning, memory retention, and in-context learning, enabling adaptive and evolving recommendation strategies. Unlike traditional systems that require explicit re-training, LLM-based agents can dynamically refine recommendations based on sequential user interactions and external contextual cues. Additionally, collaborative multi-agent systems can further enhance recommendations by enabling multiple AI agents to exchange information, reason collectively, and optimize decision-making.
    \item \textbf{Expanding Information Modalities:} RSs primarily rely on ID-based and textual information, limiting their ability to fully understand user preferences. 
    In contrast, multi-modal agentic systems can process diverse input signals, including images, audio, structured metadata, and behavioural cues, leading to richer and more context-aware recommendations. 
    % By incorporating multi-modal learning, 
    Thus, agentic systems can capture holistic user intent, bridging the gap between implicit and explicit preference signals.
    \item \textbf{Evolving User Interfaces: From Passive to Proactive Recommendation:} Traditional recommendation paradigms primarily function as passive systems, responding to user queries with static suggestions. Conversational recommenders improve engagement but still rely on user-initiated interactions. Agentic systems introduce a proactive user experience, where AI-powered multi-modal agents continuously adapt, predict user needs, and autonomously refine recommendations before explicit queries occur. This shift not only enhances user satisfaction but also opens the door for highly personalized, real-time, and contextually aware recommender systems.
    
\end{itemize}

Given these advancements, the evolution towards multi-modal LLM-driven agentic recommenders represents a promising and inevitable trajectory. These systems combine autonomy, adaptability, and multi-modal intelligence, paving the way for self-improving, memory-driven, and highly personalized recommendation experiences that surpass the capabilities of existing models.

% https://arxiv.org/pdf/2409.11393v1
\section{Formulation}

% https://tsinghua-fib-lab.github.io/AgentSocietyChallenge/pages/overview.html

An \textit{Agentic Recommender System} \cite{agent4rec, AgentCF} is a system in which agents autonomously generate personalized recommendations by interacting with users and adapting to their preferences over time. Formally, it can be defined as a tuple $(U, I, A, E, R)$, where $U$ is the set of users, $I$ is the set of items, $A$ is the set of agents, $E$ is the set of environmental contexts and $R: U \times E \times A \rightarrow P(I)$ is the recommendation function that maps users, contexts, and agents to a probability distribution over items $P(I)$. Each agent $a \in A$ operates autonomously by perceiving the state $s = f(u, e)$, making decisions based on its policy $\pi_a(s)$, and learning from user feedback to optimize an objective function, maximizing expected user utility:
\begin{equation}
    \max_{\pi_a} \, \mathbb{E} \left[ U\left( u, R(u, e, a) \right) \mid \pi_a \right].
\end{equation}
The key characteristics of such a system include autonomy, adaptability, and enabling agents to provide dynamic and personalized recommendations through continuous learning and user engagement.
To illustrate our formulation of 
% the components in 
the architecture of agentic recommender systems, we present the notation table in Table \ref{notation_table}.

\subsection{The User Profiling module:} 
The User Profiling Module is dedicated to constructing comprehensive profiles, such as behaviours for each user. The function can be define as $ P: U \times T \rightarrow \mathcal{S} $, where $P(u, t)$ represents the evolving profile of user $u$ at time $t$. This profile is dynamically updated based on historical interactions $ H(u, t) $, contextual features $ C(u, t) $, and external signals $ X(u, t) $, modeled as:
\begin{equation}
    P(u, t) = f(H(u, t), C(u, t), X(u, t); \theta_P).
\end{equation}
To adapt to new user behaviours,  profile updates incrementally as:
\begin{equation}
    P(u, t+1) = P(u, t) + \eta \cdot \Delta P(u, t),
\end{equation}
where \( \Delta P(u, t) \) represents changes based on recent interactions, and \( \eta \) controls the update rate.

The user profiling module employs machine learning techniques to adaptively refine user profiles over time. It synthesizes information from diverse sources and external contextual signals, to create a multidimensional view of the user's preferences. For instance, RecAgent \cite{wang2023recagent} utilizes large language model-based agents to simulate user behavior and refine profiling accuracy. Additionally, Rec4Agentverse \cite{zhang2024} leverages large language model-based agents for prospect personalized recommendations, allowing for finer-grained user representations.

In contemporary practice, profiling modules also leverage MLLMs to process unstructured data modalities, such as textual reviews and visual preferences. MACRec \cite{wang2024macrec} explores multi-agent collaboration frameworks to enhance user profiling through cooperative agent learning, ensuring robust profile evolution over time. Meanwhile, AgentCF \cite{AgentCF} integrates autonomous learning language agents to collaboratively refine user profiles, reinforcing adaptive personalization. By maintaining both static and dynamic aspects of user preferences, this module ensures the recommendations are contextually appropriate, significantly enhancing user satisfaction in the system. The integration of reinforcement learning frameworks like SUBER \cite{liu_suber} helps model long-term user behaviors by simulating future interactions to predict evolving preferences.

\subsection{The Planing module:}
The Planning Module empowers agents to formulate strategic decisions regarding which items to recommend. Using the user profiles from the User Profiling Module and considering the current environmental context $e \in E$, the module is defined as:
\begin{equation}
    s = f(u, e),
\end{equation}
where $f: U \times E \rightarrow \mathcal{S} $ maps users and contexts to a state space $ \mathcal{S} $. for each user-agent pair. This module functions as the core of the decision-making of the Agentic Recommender System, the Planning Module leverages advanced optimization techniques, such as Markov Decision Processes (MDPs) and reinforcement learning, to ensure that decisions are both rational and aligned with user objectives. Similar approaches have been explored in recent research on RSs, such as MACRec \cite{wang2024userbehaviorsimulationlarge} for multi-agent collaboration and Agent4Rec \cite{agent4rec}, which introduces generative agents for recommendation. In scenarios where user preferences conflict with immediate contextual constraints, the module employs multi-objective optimization to balance trade-offs effectively, similar to approaches used in BiLLP \cite{Shi_2024}, which frames recommendation as a long-term planning problem.

By simulating potential sequences of recommendations and user responses, the module can adjust strategies to minimize risks, predictive modeling is also emphasized in RecMind \cite{wang2024recmindlargelanguagemodel}, which integrates LLMs into sequential recommendation. Additionally, it can incorporate collaborative and competitive dynamics among agents, allowing for coordinated actions in multi-agent systems \cite{fang2024multi} or personalized prioritization in single-agent setups \cite{AgentCF}.

The Planning Module also enables hierarchical planning and ensures that each sub-recommendation aligns with the overall objective, creating a coherent and seamless user experience. Recent advancements in AI-driven recommendation, such as AutoConcierge 
\cite{zeng2023automatedinteractivedomainspecificconversational}, which focuses on interactive goal-based recommendations, supports this hierarchical approach to structured decision-making.

% TOWARDS FULL DELEGATION: DESIGNING IDEAL AGENTIC BEHAVIORS FOR TRAVEL PLANNING

\begin{table}[tp]
    \centering
    % \small
    \begin{tabular}{c|l}
        \toprule
        \textbf{Symbol} & \textbf{Description} \\ 
       \midrule
        \( U, I, A, E \) & Users, items, agents, environments \\ 
        \( R: U \times E \times A \rightarrow P(I) \) & Recommendation function \\ 
        \( s = f(u, e) \) & User state representation \\ 
        \( \pi_a(s) \) & Agent policy \\ 
        \( P(I) \) & Item distribution \\ 
        \( H(u, t) \) & User interaction history \\ 
        \( C(u, t) \) & Contextual factors \\ 
        \( X(u, t) \) & External signals \\ 
        \( P(u, t) \) & User profile \\ 
        \( M(u, t) \) & Memory function \\ 
        \( \mathcal{A}(s, a) \) & Action selection function \\ 
        \bottomrule
    \end{tabular}
    \caption{Summary of notations used in agent-based RSs.}
    \label{notation_table}
    % \vspace{-3em}
\end{table}

\subsection{The Memory module:}
The Memory Module functions as a dynamic storage system that retains historical data on user interactions and feedback. It serves as a critical component for enabling the Agentic Recommender System to build continuity and context awareness over time. Formally, it maintains a memory function $ M: U \times T \rightarrow \mathcal{M}$, where:
\begin{equation}
    M(u, t) = g(H(u, t), C(u, t); \theta_M),
\end{equation}
By storing and retrieving historical data, this module ensuring that future recommendations are informed by accumulated insights. Systems such as RecMind \cite{wang2023recmind} leverage LLMs for memory-driven recommendations, enhancing continuity in RSs.

The Memory Module is designed to support both short-term and long-term memory functionalities. Short-term memory stores recent interactions, enabling the system to adapt to immediate user needs and preferences. In contrast, long-term memory archives broader behavioural patterns, which are crucial for understanding shifts in user behaviour over time. Together, these memory layers create a holistic view of the user, balancing transient interests with persistent inclinations. Similar architectures are explored in SUBER \cite{liu_suber}, an RL-based framework that simulates human behaviour for adaptive recommendation learning. To manage large-scale data effectively, the Memory Module employs advanced data structuring techniques to utilizes efficient retrieval, often powered by neural attention models, to access relevant historical data in real-time. This capability is similar to BiLLP \cite{Shi_2024}, which positions LLMs as learnable planners to enhance long-term recommendation strategies. An essential feature of the Memory Module is its ability to integrate cross-session data. Systems like AgentCF \cite{AgentCF} incorporate collaborative learning mechanisms, enabling memory-enhanced interactions among language agents in multi-agent recommendation.

\subsection{The Action module:}
The Action Module is responsible for executing the decisions made by the Planning Module, dynamically selecting and delivering recommendations to users. Given a user $ u \in U $, an agent  $ a \in A $ , and an environmental state $ e \in E $, the system defines an action selection function $ \mathcal{A} : \mathcal{S} \times A \rightarrow P(I) $, where:
\begin{equation}
    \mathcal{A}(s, a) = \pi_a(s),
\end{equation}
where $ \pi_a(s) $ represents the agent’s policy for selecting a probability distribution over items $ P(I) $, given the current state $ s = f(u, e) $.
Modern recommender systems increasingly integrate agentic approaches that allow for interactive decision-making. For instance, Agent4Rec \cite{agent4rec} introduces generative agents that enable personalized through reinforcement learning. Similarly, RecAgent \cite{wang2023recagent} uses a simulation of user behaviour with agents based on large language models to refine recommendation strategies.

Multi-agent frameworks have been explored to facilitate collaboration and competition in recommendation settings. MACRec \cite{wang2024macrec} demonstrates the potential of multi-agent collaboration frameworks for improving recommendation diversity and accuracy. Moreover, MACRS \cite{fang2024multi} expands on this by introducing multi-agent conversational recommender systems that coordinate interactions across multiple agents to optimize recommendations in real-time.
Conversational RSs play a crucial role in the Action Module by enabling context-aware responses. 
RecLLM \cite{friedman2023leveraginglargelanguagemodels} and CSHI \cite{zhu2024llm} focus on leveraging large language models to enhance conversational interactions,
providing scalable and controllable user simulations.
% for testing recommendation strategies. 
RecMind \cite{wang2023recmind} employs large language models to power agent-based recommendations, ensuring responses are aligned with evolving user intents. 
LLM4Rerank \cite{gao2025llm4rerank} further enhances recommendation effectiveness through re-ranking mechanisms optimized by LLMs.

A novel direction is tool-augmented recommendations (\emph{e.g.}, ToolRec \cite{Zhao_2024toolrec}), which leverages tool learning to enhance recommendation accuracy and usability. 
Similarly, RAH \cite{shu2023rah} presents a human-centered framework that balances LLM-powered recommendations with human oversight improving user satisfaction.

\section{Key Research Questions in LLM-ARS}\label{sec:RQs}
After formulating an agentic recommender system and examining its key components, the next step is to address fundamental challenges in integrating LLM-driven agentic capabilities. These challenges span reasoning, user modeling, multimodal fusion, lifelong personalization, decision-making frameworks, controllability, and so on.
% As RS increasingly integrates LLM-driven agentic capabilities, they must address fundamental challenges from both agent and RS perspectives. 
% such as user modeling, multimodal fusion, lifelong personalization, decision-making frameworks, controllability, and evaluation paradigms.
To systematically analyze these challenges and explore novel solutions, we structure our discussion around the following key research questions (RQs).

\textbf{RQ1:} How can LLM-based agents benefit recommender systems through reasoning, planning, and collaboration?

\textbf{RQ2:} How can agentic recommender systems effectively leverage (M)LLM to improve user understanding and decision-making? 
% (user simulation/ recommender systems)

\textbf{RQ3:} What novel architectures or learning paradigms are needed to enable agentic RSs? 
% (single/multi-agent framework)

\textbf{RQ4:} What are the key challenges in integrating agentic decision-making and multimodal reasoning into RSs?
% (open problem: multi-modal fusion, efficiency)

\textbf{RQ5:} How can we evaluate the effectiveness and robustness of agentic recommender systems powered by multimodal LLMs? 
% (open problem:  benchmark, offline/online evaluation)

\textbf{RQ6:} How can agentic recommender systems balance autonomy and controllability while utilizing MLLMs? 
% (open problem: hallucination, bias, explainable)

\textbf{RQ7:} How can agentic recommender systems achieve life-long personalization while mitigating catastrophic forgetting?
% and adapting to evolving user preferences? 

\section{LLM-based Agentic Reasoning, Planning, and Collaboration (RQ1)}\label{sec:agent_perspective}
% From the perspective of LLM agents, recommendation poses challenges in long-term planning and reasoning over personalized contexts and preference feedback.
% Different from conventional recommendation methodologies, 
% which learn from historical data to capture the statistical patterns of user behaviours \cite{wang2019sequential}, user-item collaborative information \cite{schafer2007collaborative}, and item-item similarity \cite{rendle2011fast},
% LLM agents enable reasoning by analyzing the contextual information of items \cite{AgentCF,wu2024coral} and the semantic information of user-item interactions \cite{wu2024coral}.
% To further explore users' long-term preferences, LLM agents enable planning to generate proactive strategies, which rely on LLM's chain-of-thought generation abilities \cite{wang2024rdrec,zhao2024lane,wu2024coral}.

% On the other hand, since LLMs are designed as general-purpose models, adapting to personalized contexts or user feedback can be challenging.
% To better simulate various users' personalities, LLM agents are enabled to roleplay through prompting \cite{AgentCF} and user modelling \cite{zhang2024llm}.
% By incorporating such LLM roleplaying agents in interactive settings, 
% the recommendation agents are further self-improved through multi-agent alignment \cite{wang2025large,wang2024reinforcement,wu2024coral}.

In this section, we explore how LLM agents face challenges in long-term planning and reasoning over personalized contexts and feedback (\textbf{RQ1}). 
Unlike conventional recommendation methods that learn from historical data to capture statistical patterns of user behavior \cite{wang2019sequential,schafer2007collaborative,rendle2011fast}, 
LLM agents analyze the contextual information of items and the semantic details of user-item interactions \cite{AgentCF,wu2024coral}. 
They further plan proactive strategies to explore long-term preferences using chain-of-thought generation \cite{wang2024rdrec,zhao2024lane,wu2024coral}.
However, as general-purpose models, LLMs find it challenging to adapt to personalized contexts or user feedback. 
To simulate diverse personalities, LLM agents roleplay via prompting \cite{AgentCF} and user modelling \cite{zhang2024llm}, and they self-improve in interactive settings through multi-agent alignment \cite{wang2025large,wang2024reinforcement,wu2024coral}.

% \subsection{Agent Architectures for RS}
% \cite{park2025agentrec} AgentRec: Agent Recommendation Using Sentence Embeddings Aligned to Human Feedback
% % Item-Language Model for Conversational Recommendation

\subsection{Planning and Reasoning in Agentic RS}

LLM agent planning in recommender systems leverages the complex reasoning and decision-making capabilities of large language models to 
decompose the recommendation process into subtasks and assign them to multiple agents for collaboration across agents.
To manage complex recommendation tasks, \citet{wang2024macrec} and \citet{fang2024multi} propose multi-agent frameworks that decompose the overall task into specialized roles,
while \citet{wang2024macrec} introduces agentic protocols including Manager, User/Item Analyst, Reflector, Searcher, and Task Interpreter.
\citet{fang2024multi} focuses on goal-oriented dialogue planning and incorporates a user feedback-aware reflection mechanism to control the conversation flow. 
To mitigate issues such as hallucinations and misalignment between semantics and behaviours, \citet{zhao2024let} employs tool learning with surrogate users and attribute-oriented tools (i.e., rank and retrieval tools), 
while \cite{li2024incorporating} integrates external knowledge and goal guidance to better reasoning grounding and proactive responses. 
To further enable exploration in planning \citet{wang2025large} develops LLM-driven policy exploration by pre-training policies with user preference distillation for deploying adaptive fine-tuning strategies.

LLM agent equips recommender systems with the reasoning capabilities of large language models to discover complex user-item relationships and 
generate interpretable and semantically meaningful recommendations. 
By further integrating structured external knowledge, distilled rationales, and memory mechanisms, LLM-based agentic frames are enabled with more contextually grounded reasoning 
while understanding various personalized behaviours and preferences in recommendation tasks.
To uncover complex user-item relationships, \citet{guo2024knowledge} leverages knowledge graphs to inject explicit relational paths into language agents, 
while \citet{wang2024rdrec} distils underlying rationales from user reviews to enrich user profiles and item contexts, which improves LLM agents' understanding of complex user-item interactions. 
To further understand the sequential context and user behaviours in conversational recommendations, \citet{xi2024memocrs} introduces memory-enhanced LLMs to track historical dialogue beliefs, 
improving on the approaches that only consider current interactions. 
To ensure explanations are both persuasive and credible, \citet{qin2024beyond} develops a credibility-aware strategy that refines outputs through self-reflection.
Focusing on the alignment of LLM reasoning with recommendation logic, \citet{zhao2024lane} proposes a non-tuning logic alignment framework using semantic embeddings and chain-of-thought prompting, 
whereas \citet{wu2024coral} augments LLMs with collaborative retrieval to ground reasoning in user-item interaction patterns.

% \subsection{Limitations} Despite promising advances, these approaches face limitations such as the absence of a unified framework that fully integrates multi-agent planning with reinforcement learning and user behavior modeling, as observed in \cite{wang2024macrec} and \cite{wang2025large}. Furthermore, improvements in robust external knowledge infusion and comprehensive dialogue management—as suggested by \cite{li2024incorporating} and \cite{fang2024multi}—remain underexplored, indicating potential avenues for future research.
% Despite these promising approaches, current methods often rely on explicit external structures—such as knowledge graphs in \cite{guo2024knowledge} or curated rationales in \cite{wang2024rdrec}—which may limit their adaptability across diverse recommendation scenarios. Moreover, while techniques in \cite{xi2024memocrs} and \cite{qin2024beyond} enhance sequential reasoning and explanation credibility, and \cite{zhao2024lane} and \cite{wu2024coral} improve logic alignment and collaborative retrieval, an integrated framework that seamlessly combines these reasoning enhancements while managing prompt capacity and computational efficiency remains underexplored.
Despite promising advances in LLM agents for planning and reasoning in recommender systems, current approaches face notable challenges. 
Methods dependent on explicit external structures—such as knowledge graphs \cite{guo2024knowledge} or curated rationales \cite{wang2024rdrec} are limited in generalizability across various scenarios. 
Although techniques in \cite{xi2024memocrs} and \cite{qin2024beyond} improve sequential reasoning and explanation credibility, 
and \cite{zhao2024lane} and \cite{wu2024coral} enhance logic alignment and collaborative retrieval, an integrated framework that aligns multi-agent reinforcement learning and planning with user behaviour modelling \cite{wang2024macrec, wang2025large} is still lacking. 

\subsection{LLM-Agent Roleplaying in User Modeling}

The exploration of LLM-agent roleplaying techniques is demanding for realistic user modelling in recommender systems, where user agents or simulators emulate human-like behaviours to capture both explicit and implicit user preferences. 
Intuitively, these methods leverage roleplay to bridge the gap between language understanding and behaviour simulation, enabling more realistic multi-agent interactions for personalized preference alignment and more rigorous evaluation.
One prominent challenge is simulating socially dynamic user-item interactions inherent in human behaviour. 
\citet{zhang2024agentcf} tackles this by simulating a collaborative learning environment where both users and items are modelled as autonomous roleplaying agents, 
thus enabling bidirectional interaction and reflective adjustment. 
In addition, \citet{wang2024user} introduces a sandbox environment where roleplaying agents are equipped with profile, memory, and action modules that interact through one-to-one and broadcast communications, 
effectively modelling social influence and conformity.
In contrast, \citet{zhang2024llm} emphasizes explicit user modelling by integrating logical reasoning with statistical insights to simulate user engagement.

Addressing the need for controllability and scalability in conversational settings, \citet{zhu2024llm} proposes a framework that utilizes roleplay to customize user simulations in real time, 
enhancing the fidelity of user modelling in conversational recommender systems. 
Additionally, to overcome limitations related to data scarcity and evaluation reliability, \cite{corecco2024llm} and \cite{ebrat2024lusifer} construct synthetic environments using LLMs as roleplaying users, 
while \cite{kim2024stop} introduces a target-free roleplay strategy to avoid bias in preference elicitation.
However, current LLM-agent roleplaying approaches in user modelling still struggle with the interpretability of simulation processes and 
% the challenge of 
capturing the complexity of human decision-making. 
Future research should focus on developing more interpretable roleplay strategies and integrating richer, multimodal behavioural data to further enhance the adaptability and realism of user modeling frameworks.

\subsection{Interaction Between Agents and Users}

LLM-based agentic recommendation systems have motivated exploring methods that enhance the realistic interaction between agents and users. 
Intuitively, these approaches leverage agent role-playing and collaborative mechanisms to bridge the gap between language understanding and complex behavioural interactions.
One of the major challenges is simulating realistic user-agent interactions by capturing both explicit semantic and implicit behaviour signals.
\citet{zhang2024agentcf} addresses this by modelling non-verbal signals (e.g., item clicking) via collaborative learning between user and item agents, 
in contrast to dialogue-centric approaches such as \cite{fang2024multi}. 
\citet{kim2024stop} further emphasizes a target-free user simulation protocol that avoids the target bias in such interactions.

Another challenge lies in integrating task-specific recommendation dynamics with interactive capabilities.
While \citet{huang2023recommender} leverages LLMs as a central controller augmented by recommendation models to enable seamless interaction, 
\citet{wang2024enhancing} focuses on enhancing high-order interaction awareness through whole-word embedding techniques. 
In multi-agent systems, collaboration in achieving effective interaction is proposed by \cite{wang2024macrec}, 
which designs specialized agents for various subtasks, 
whereas \cite{fang2024multi} suggests feedback-aware reflection for controlled dialogue flow. 
However, existing works still fall short in robustly modelling the dynamic evolution and collaborative evolution of extended agent-user interaction, fully integrating adaptive feedback mechanisms. 
Future research should explore strategies for multi-agent planning and reasoning to align dynamic user-item interaction.
% , enhancing the fidelity and adaptability of interactive systems.

% human value alignment
%RosePO: Aligning LLM-based Recommenders with Human Values

\subsection{ Agent Self-improvement}

Finally, we discuss how agents can further evolve and self-improve in a recommendation environment by continuously incorporating rich interaction signals. 
Leveraging large language models (LLMs) to simulate and distil these interactions, recent approaches aim to bridge the gap between static offline training and evolving online deployment.
Synthesizing effective feedback from sparse data can significantly scale up the offline training of LLM agents.
\citet{wu2024coral} integrates collaborative information to enrich the interaction context, in addition to the approach \cite{wang2024reinforcement} that directly generates feedback via LLM capabilities. 
Addressing the challenge of distribution shift and limited exploration in offline reinforcement learning, 
\citet{wang2025large} introduces an Interaction-Augmented Learned Policy (iALP) that pre-trains policies with distilled user interaction data augmented by LLMs, 
while \citet{wang2024reinforcement} employs an LLM as an environment to verbally model states and rewards from real interaction feedback. 
Meanwhile, in the domain of adaptive agent selection, \cite{park2025agentrec} leverages sentence embeddings aligned with human feedback to recommend the most appropriate agent based on interactive prompting, ensuring adaptability in dynamic settings. 
Confronting the need for explainability in self-improvement, \cite{zhao2024lane} proposes a logic alignment strategy that enables LLM reasoning in online systems, providing interpretable recommendations grounded in explicit interaction semantics.
However, current methods are still limited in the reliance on synthetic or simulated interaction data, which may not fully capture the complexities of real-world environments. 
In addition, the sim-to-real gap can be additionally challenging, which requires robust offline policy evaluation, and smart online adaptation strategies.

% \subsection{Agentic Approaches to User Modeling(User profiling)}

% \section{from RS perspective}\label{sec:RS_perspective}

% (open problem: life-long personalization)

% From the perspective of the recommender system field.
% In the LLM-powered autonomous agent system, LLMs function as the agent’s brain, complemented by key components like planning, memory, and tool use~\cite{weng2023prompt}. 
% There are many inspiring works such as AutoGPT and BabyAGI proving the LLM-based agents' potential, These agents can store their past experiences and make better decisions for future behaviors.
% In Recommender System scenarios, agents are typically represented as either User Simulators or the Recommender System itself as shown in Figure \ref{fig:agent}.

\section{LLM Agents for Enhanced User Understanding and Decision-Making (RQ2)}\label{sec:RS_perspective}

\begin{figure*}[t]  
\small
	\centering
\includegraphics[width=1\linewidth]{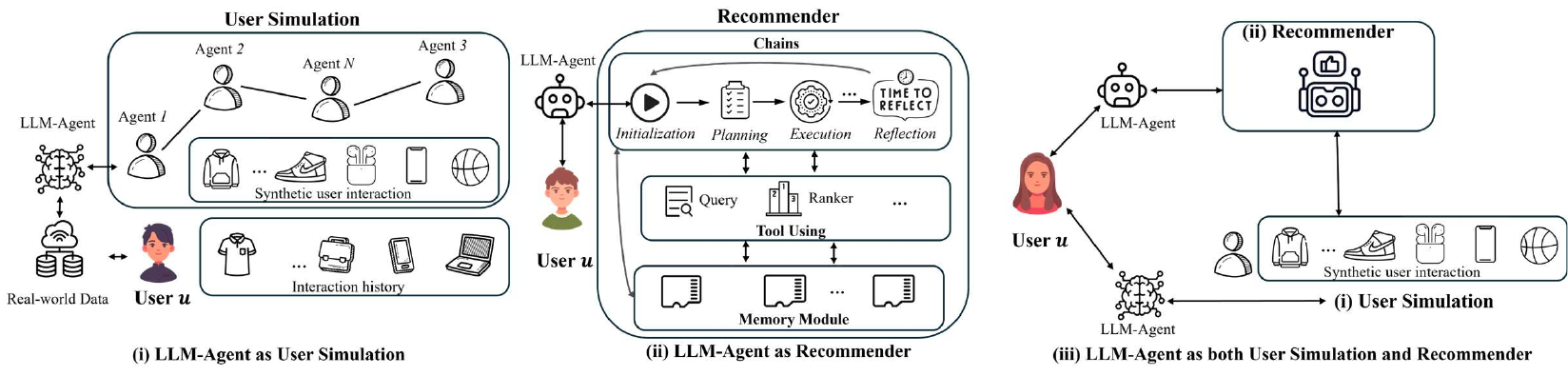}
\caption{Different types of personalized LLM-based agents in LLM-ARS, where (i) LLM-Agent simulates user behavior, (ii) LLM-Agent acts as a recommender, and (iii) LLM-Agent functions as both user simulation and recommender.}
	\label{fig:agent}
\end{figure*}

% \textbf{RQ2:} How can agentic recommender systems effectively leverage (M)LLM to improve user understanding and decision-making? 
From the perspective of the RS field, LLM-powered autonomous agent systems position LLMs as the core "brain" of the agent, supported by essential components such as planning, memory, and tool utilization \cite{weng2023prompt}. Prominent works like AutoGPT and BabyAGI have demonstrated the immense potential of LLM-based agents, particularly in their ability to store past experiences and leverage them to make more informed decisions (\textbf{RQ2}). In RS scenarios, these agents are often conceptualized as user simulators or the RS itself, as illustrated in Figure \ref{fig:agent}.

% \begin{figure*}[t]  
% \small
% 	\centering
% \includegraphics[width=1\linewidth]{figures/agent.pdf}
% \vspace{-10pt}
% \caption{Different types of personalized LLM-based agents in LLM-ARS, where (i) LLM-Agent simulates user behavior, (ii) LLM-Agent acts as a recommender, and (iii) LLM-Agent functions as both user simulation and recommender.}
% % : (a) Agent as User Simulator and (b) Agent as Recommender System (c) as both User Simulator and RS.}
% 	\label{fig:agent}
% \end{figure*}
% \setlength{\abovecaptionskip}{-2pt}  % 控制标题上方的间距

\subsection{User Simulation in LLM-ARS}

Simulating user behaviors is essential for training large-scale RSs, given the challenges of data scarcity, ethical concerns, and cold-start issues in real-world interaction data. Traditional methods \cite{ZhuLLWGLC17,RecSim} struggle to model complex and evolving user behaviors, while recent advances in LLMs provide a promising alternative by enabling more adaptive and realistic simulations.

% \cite{wang2023recagent}.
Most works leverage LLM-powered personalized agents to emulate user interactions. RecAgent \cite{wang2023recagent} treats each user as an autonomous agent capable of interacting freely within a simulated environment, capturing both conventional RS behaviors such as browsing and clicking, as well as external influences like social interactions. Extending this idea, Agent4Rec \cite{agent4rec} simulates 1,000 generative agents in a movie RS, where users engage with recommendations in a page-by-page manner, taking diverse actions that better approximate real-world decision-making.
Beyond individual user agents, collaborative simulation frameworks have emerged to model multi-agent dynamics. LLM-InS \cite{Feiran24} predicts user interactions with cold-start items, simulating clicks from a subset of recalled users to generate synthetic interactions that update item embeddings. Zhang \textit{et al.} \cite{zhang2024llm} integrate LLM-based logical reasoning with statistical modeling, extracting user preferences from item characteristics and engagement history to improve the fidelity of simulated behaviors.
AgentCF \cite{AgentCF} extends the paradigm by treating both users and items as interactive agents, fostering a co-evolutionary learning process that optimizes user-item interactions. USimAgent \cite{Erhan24} focuses on search behavior simulation, capturing querying, clicking, and stopping behaviors to generate realistic search task interactions. BASES \cite{BASES24} scales this concept further, utilizing LLM-based agents to create large-scale user profiles and diverse search behaviors across multiple linguistic benchmarks.

Despite advancements, LLM-driven simulators face critical limitations. Many rely on predefined heuristics or scripted rules, failing to capture emergent or long-term behavioral patterns. While LLMs approximate user preferences, they lack the ability to model cognitive biases, evolving interests, or contextual decision-making shifts. Scalability is also a concern: synthetic interactions can be generated at scale, but their real-world validity remains uncertain, and over-reliance on simulated data risks introducing biases. Future work should focus on adaptive, feedback-driven frameworks that integrate real-world behavioral signals, refine user modeling beyond static preferences, and establish validation mechanisms for LLM-generated interactions in RS applications.

\subsection{Improving Personalized Recommendations with LLM-driven Decision-Making}

Leveraging the advanced reasoning, reflection, and tool-usage capabilities of LLM agents, recent approaches explore their role as decision-making agents to enhance personalized recommendations. Unlike level 0-2 RS models, LLM-ARSs dynamically adapt to user needs by integrating planning, self-reflection, and external tool interactions.
% leveraging the robust capability of LLMs, including reasoning, reflection, and tool usage for recommendation.
The RAH framework \cite{Yubo23}, incorporating LLM-based agents and a Learn-Act-Critic loop, improve alignment with user personalities and mitigate biases. Then, Wang \textit{et al.} \cite{wang2023recmind} first introduces a Self-Inspiring planning algorithm that keeps track of all past steps of the agent to help generate new states. At each step, the agent looks back at all the paths it has taken before to figure out what to do next. This approach aids in employing databases, search engines, and summarization tools, combined with user data, for producing tailored recommendations. InteRecAgent \cite{InteRecAgent} model the LLMs as the brain, while recommendation models serve as tools that supply domain-specific knowledge, then LLMs can parse user intent and generate responses. They specify a core set of tools essential for RS tasks—Information Query, Item Retrieval, and Item Ranking—and introduce a candidate memory bus, allowing previous tools to access and modify the pool of item candidates.

However, key challenges remain, such as ensuring long-term consistency in recommendations, balancing LLM-ARS generalization with domain-specific accuracy, and mitigating potential biases introduced by LLM-generated reasoning. Future research should focus on integrating user feedback loops, enhancing interpretability, and optimizing the efficiency of tool-augmented LLM decision-making to fully realize the potential of LLM-ARS.

\section{Framework and Learning Paradigms (RQ3)} 
%for ARS (RQ2)}

To enable LLM-ARS, novel frameworks and learning paradigms are required to enhance autonomy, adaptability, and human alignment (\textbf{RQ3}). We categorize these advancements into three key areas: single-agent architectures, which focus on individual agents as decision-makers; multi-agent collaboration, which leverages interactions among multiple agents to improve reasoning and adaptability; and human-LLM hybrid architectures, which emphasize collaboration between human users and LLM-based agents to refine personalization, control, and interpretability in recommendations.

\textbf{Single-Agent Framework for RS:} LLM-powered single-agent frameworks enable autonomous decision-making in RSs by integrating reasoning, memory, and planning. The RAH framework \cite{Yubo23} employs a Learn-Act-Critic loop to iteratively refine recommendations, improving personalization and reducing bias. Wang \textit{et al.} \cite{wang2023recmind} introduce Self-Inspiring Planning, where an LLM agent retrospectively analyzes past decisions to optimize future choices while leveraging external tools like search engines and summarization models. InteRecAgent \cite{InteRecAgent} further enhances this paradigm by treating LLMs as decision-making cores, selectively invoking domain-specific tools (e.g., retrieval and ranking modules) and maintaining long-term candidate memory for adaptive ranking. These architectures transform LLMs from passive generators into adaptive decision-makers, enabling more context-aware, interactive recommendations. However, they face scalability challenges and lack collaborative reasoning in multi-domain scenarios.

\textbf{Multi-Agent Framework for RS}
% Collaborative Reasoning:} 
Multi-agent frameworks extend single-agent frameworks by incorporating specialized agents that communicate and collaborate to enhance decision-making. Instead of relying on a single agent for all tasks, these frameworks assign distinct roles to different agents, enabling parallelized reasoning, task specialization, and self-organizing interactions. Wang \textit{et al.} \cite{Zhefan24} propose MACRec, where agents such as a Manager, Analyst, and Reflector collaborate on tasks like rating prediction, sequential recommendation, and explanation generation, improving adaptability and interpretability. PUMA \cite{Hongru} further integrates a shared memory system, allowing agents to retrieve past interactions for enhanced personalization. Compared to single-agent models, multi-agent frameworks offer better scalability, modularity, and reasoning efficiency, yet face challenges in coordination, redundancy reduction, and consistency maintenance across interacting agents.
% To address gaps in multi-agent collaboration for recommendation, Wang \textit{et al.}~\cite{Zhefan24} introduce MACRec, a hierarchical multi-agent framework where specialized agents—including Manager, User/Item Analyst, Reflector, Searcher, and Task Interpreter—coordinate to tackle various recommendation tasks such as rating prediction, sequential recommendation, conversational recommendation, and explanation generation. By distributing responsibilities, MACRec enhances adaptability and reduces reliance on a single LLM model, making the system more interpretable and modular.
% Similarly, PUMA~\cite{Hongru} introduces a memory-augmented multi-agent system, where agents can retrieve and reuse past user interactions to improve recommendation alignment. By incorporating a shared memory system, PUMA enables agents to refine predictions based on historical behavior, improving long-term personalization.
% Compared to single-agent systems, multi-agent architectures improve scalability, enable better reasoning through division of labor, and allow for modular improvements in specific sub-tasks. However, challenges remain in efficiently coordinating agents, preventing redundant computations, and ensuring consistency in decision-making across multiple interacting agents.

\textbf{Human-LLM Hybrid Framework for RS:} While LLM-powered agents enhance automation, human-in-the-loop architectures are crucial for improving interpretability and 
%personalzation
fairness in RSs. Recent works explore collaborative frameworks where user feedback guides LLM-driven reasoning, ensuring transparency and control.
Shu \textit{et al.} \cite{ShuZGZLLG24} propose 
% RAH (Recommender system, Assistant, and Human), where an 
the LLM-powered assistant mediates between users and RSs. Using a Learn-Act-Critic loop with built-in reflection, the assistant refines recommendations by resolving preference inconsistencies. It also incorporates privacy-preserving mechanisms, allowing users to filter content and adjust recommendations dynamically. Beyond direct interaction, hybrid frameworks embed user intent into LLM-based reasoning. Ning \textit{et al.} \cite{Ning2024} integrate user embeddings with LLMs via a pretrained encoder and cross-attention, capturing long-term preferences more effectively. Shao \textit{et al.} \cite{shao2024ulmrec} further bridge the semantic gap between LLM reasoning and structured user data through vector quantization and preference alignment. To formalize design principles for human-centered agentic RSs, Deng \textit{et al.} \cite{DengLZYC24} introduce a taxonomy spanning Intelligence, Adaptivity, and Civility, providing guidelines to develop ethically adaptive, user-aligned conversational recommenders. 
In summary, single-agent systems enable autonomous reasoning and memory integration, while multi-agent architectures enhance collaboration and modularity. 
Human-LLM hybrids further improve interpretability and personalization. 
Key challenges include balancing autonomy with user control, optimizing coordination, and mitigating biases while ensuring generalization. 
Future research should develop adaptive architectures that unify reasoning, collaboration, and user alignment for fully interactive, context-aware systems.

\section{Open Problems and Opportunities} 
% in LLM-ARS}
% In this section, we discuss key open problems and opportunities for advancing LLM-ARS (RQ4-RQ6).

% focusing on efficiency, benchmarking, personalization, safety, and beyond.

\subsection{Multimodal Reasoning in LLM-ARS (RQ4)}
In this section, we investigate key challenges in integrating agentic decision-making and multimodal reasoning into RSs (\textbf{RQ4}).

\par \textbf{Multimodal Fusion:} Multimodal fusion is crucial for agentic RSs integrating multiple LLMs and tools, yet it remains challenging. Potential strategies include encoder-decoder, attention, GNN, and generative neural network (GenNN)-based fusion. Encoder-decoder models unify multimodal features in a shared space for task-specific decoding \cite{tan2022efficient, khattar2019mvae}, while attention-based fusion enhances cross-modal dependencies \cite{wu2021multimodal, lu2022multimodal}. GNN-based approaches jointly model structured and unstructured data \cite{qian2021knowledge, tao2020mgat}, and GenNN-based fusion synthesizes modalities while handling missing data \cite{sahu2019adaptive}. Effective fusion strengthens reasoning and factual grounding, ensuring robust decision-making in LLM-ARS.

\par \textbf{Multimodal Reasoning:} Aligning (M)LLM commonsense reasoning with recommendation tasks remains a key challenge. While (M)LLMs excel in open-domain reasoning, they often lack the task-specific adaptability needed for user preference modeling and sequential decision-making. Their reasoning is optimized for general understanding rather than multimodal user intent inference, leading to inconsistencies in recommendation relevance. Addressing this requires fine-tuning with domain-specific constraints, integrating structured knowledge, and optimizing reasoning for personalized decision-making in multimodal contexts.

\par \textbf{Efficiency:}
Efficiency remains a critical challenge for LLM-ARS, especially as they orchestrate multiple specialized tools or models. Current RSs often incur significant computational overhead when integrating LLMs with external APIs for multimodal tasks, leading to latency issues. Optimizing the agent pipeline for speed and resource utilization while maintaining accuracy is essential. Promising directions include developing lightweight agents, reducing redundant computations through shared intermediate outputs, and exploring model compression techniques for LLMs within agents.

% (Model distillation)
% Adaptive computation, such as activating specific modules on demand, is another promising avenue.

\subsection{Benchmarking of LLM-ARS (RQ5)}

Benchmarking LLM-ARS presents unique challenges beyond established metrics for LLMs and standalone RSs (\textbf{RQ5}). Comprehensive frameworks like AgentBench \cite{agentbench} are essential for assessing multi-turn interaction quality, cross-modal effectiveness, and adaptability to user feedback. Effective evaluation demands standardized datasets and protocols that capture real-world complexity, including dynamic personalization and multimodal workflows. Robust assessment should integrate qualitative insights with quantitative metrics, measuring coherence, responsiveness, and contextual relevance under evolving conditions. 
Stress-testing adaptability to emergent feedback ensures sustained performance. 
Developing realistic simulation environments aligned with real-world use cases will enhance benchmarking transparency and drive iterative improvements in ARS.
% \par \textbf{Benchmarking:}
% % agentbench
% While LLMs and standalone RSs have established evaluation metrics, benchmarking agentic systems introduces unique challenges. Comprehensive frameworks, such as \textit{AgentBench} \cite{agentbench}, are needed to evaluate multi-turn interaction quality, cross-modality effectiveness, and adaptability to user feedback. Establishing standardized datasets and evaluation protocols that reflect real-world complexity (e.g., dynamic personalization, multimodal workflows) is crucial for advancing research in ARS.

% \par \textbf{Evaluation:} Evaluating the effectiveness and robustness of LLM-ARS demands an integrated framework that merges qualitative insights with rigorous quantitative metrics. It is crucial to design evaluation protocols that not only measure multi-turn interaction coherence and cross-modal performance but also capture the dynamic nuances of personalized user experiences in real-world scenarios. Robust assessment should extend to stress-testing the system’s adaptability to evolving feedback and emergent contexts, ensuring that recommendations remain contextually relevant under varying operational conditions. This necessitates the development and adoption of standardized datasets and simulation environments that accurately mirror the complexity of multimodal workflows and user behaviours. By aligning evaluation strategies with authentic use cases and continuous learning cycles, researchers can systematically benchmark system performance, foster transparency, and drive iterative improvements in the design of ARS.

\subsection{Balancing Autonomy and Controllability in LLM-ARS (RQ6)}

Ensuring a balance between autonomy and controllability in LLM-ARS requires addressing key challenges such as hallucination, explainability, and safety (\textbf{RQ6}). While agentic RSs benefit from LLMs’ ability to generate flexible and adaptive recommendations, uncontrolled generation can lead to unrealistic, irrelevant, or even harmful recommendations. Below, we discuss how these challenges manifest in RS scenarios and the strategies to mitigate them.

\par \textbf{Hallucination:}
Hallucination in LLM-ARSs commonly occurs when generated items fall outside the valid item pool (OOV items) or when the model fabricates user preferences inconsistent with real behavior. This issue arises from LLMs’ open-ended generative nature.  This issue arises because LLMs, unlike retrieval-based RSs, do not inherently constrain outputs to an existing catalog. For instance, an LLM might recommend an out-of-vocabulary (OOV) item that does not exist in the system’s database, generate unrealistic item-attribute pairings in multimodal RSs, or infer user interests based on semantic associations rather than actual interactions. Such errors are especially problematic in domains like e-commerce, where recommending unavailable products could degrade user trust. To mitigate hallucination, several strategies have been proposed. Database-grounded generation techniques ensure that LLMs reference an external item pool before finalizing recommendations \cite{zhao2024effectively}. Reflective instruction tuning helps refine constraints on generation \cite{zhang2024reflective}, while hallucination detection frameworks flag outputs that lack factual grounding \cite{yu2024hallucidoctor}. At inference time, methods such as adaptive grounding \cite{chen2024halc} and self-introspective decoding \cite{huo2024self} validate recommendation outputs in real-time, ensuring that generated suggestions align with available content. By applying these techniques, LLM-ARSs can maintain generative flexibility while preventing misleading recommendations.

\par \textbf{Explainability and Trust:} Ensuring explainability and user trust is a key challenge in LLM-ARS, as LLM-driven models often function as opaque decision-makers. Unlike traditional RSs with structured optimization criteria, LLM-ARS recommenders rely on implicit reasoning, making it difficult to trace their decisions. This opacity can lead to skepticism, especially when recommendations seem arbitrary or inconsistent. For instance, an LLM in a conversational RS might suggest a book based on inferred emotional tone rather than explicit preferences, while a multimodal RS may recommend a movie based on textual reviews without justifying it through content features like genre or cast. To improve transparency, recent methods explore natural language rationale generation \cite{chen2021generate}, structured decision paths via external knowledge graphs \cite{xian2019reinforcement, lyu2022knowledge}, and cross-attention mechanisms that embed user interactions into LLM reasoning \cite{li2023personalized}. Chain-of-thought prompting further enhances interpretability by breaking down recommendations step by step \cite{li2023prompt}. Aligning model reasoning with explicit knowledge sources strengthens user trust and control over recommendations.

\par \textbf{Safety and Vulnerability:} 
% As LLM-ARSs become more autonomous, ensuring safety and robustness is paramount \cite{he2024security}. Key challenges include preventing adversarial attacks (e.g., prompt injection, poisoned datasets) and mitigating unintended consequences from agentic decisions (e.g., biased or harmful recommendations) \cite{zhan2024injecagent,zeng2024autodefense,hua2024trustagent,xiang2024guardagent}. Future research should focus on developing robust agent architectures, integrating adversarial training techniques, and implementing rigorous auditing systems. Designing agents that can gracefully handle unexpected user inputs or environmental changes without compromising system integrity is essential. Establishing ethical guidelines for agentic behavior and incorporating explainability to build trust are critical steps \cite{deng2024ai}.
As LLM-ARSs become more autonomous, ensuring safety and robustness is critical, particularly in preventing adversarial manipulation and unintended biases. Malicious users can exploit vulnerabilities through prompt injection, data poisoning, and adversarial attacks, leading to biased or harmful recommendations \cite{zhan2024injecagent, zeng2024autodefense}. Additionally, LLM-based RSs risk reinforcing historical biases, over-optimizing for engagement at the cost of diversity and fairness. Over-personalization further exacerbates filter bubbles, limiting content discovery.
Addressing these risks requires multi-layered safeguards. Adversarial training enhances resilience \cite{xiang2024guardagent}, while fairness-aware algorithms impose constraints to mitigate bias \cite{hua2024trustagent}. User feedback loops enable manual overrides, preserving user agency. Governance frameworks establish ethical boundaries for autonomous recommenders \cite{deng2024ai}. Together, these mechanisms strengthen the security and reliability of LLM-ARS, ensuring autonomy aligns with ethical responsibility.

\subsection{Life-long Personalization in LLM-ARS (RQ7)}
% \subsection{Life-long Personalization}
Personalization in agentic recommender systems is currently limited to short-term memory or static user profiles \cite{Tiannan2024}. Life-long personalization introduces the concept of continual learning, where agents evolve with the users' preferences over time (\textbf{RQ7}). Rather than passively generating recommendations, these agents should actively engage with users, clarify ambiguities, and refine their understanding through long-term feedback loops. 
Challenges include handling catastrophic forgetting, aligning learning with changing user preferences, and maintaining scalability as user interaction histories grow. 
Approaches such as meta-learning, episodic memory systems, and AI personas—persistent representations \cite{Wen0SLT0CT24} of user preferences—can provide promising solutions. These approaches ensure that agents adapt to users’ evolving needs across diverse contexts and applications.

% continual learning
% AI PERSONA:Towards Life-long Personalization of LLMs
%LIBER: Lifelong User Behavior Modeling Based on Large Language Models

\section{Conclusion}

This perspective paper first examines the integration of LLMs into agentic RSs, highlighting their role in enabling dynamic, adaptive, and multimodal interactions. We categorize recent advancements into single-agent, multi-agent, and human-LLM hybrid architectures, analyzing their impact on personalization, transparency, and reasoning. Despite these advancements, challenges such as efficiency, hallucination, safety, and lifelong learning remain critical. To address these, we outline future directions, including scalable architectures, robust evaluation frameworks, and improved domain generalization. As agentic RSs evolve, ensuring a balance between autonomy and controllability will be essential for building trustworthy, context-aware, and ethically aligned recommender systems.

% This perspective paper investigated Agentic Recommender Systems in the Era of Multimodal Large Language Models

% The integration of LLMs into agentic systems has opened new horizons for RSs, enabling more dynamic, adaptive, and multimodal interactions. This perspective paper provides a perspective view of recent advancements, highlighting the synergistic relationship between LLMs and agents in addressing key challenges such as personalization, transparency, and reasoning across complex tasks. However, as we explore the capabilities of these systems, critical challenges such as efficiency, hallucination, safety, and life-long learning remain open problems. Addressing these issues will require the development of scalable architectures, robust evaluation frameworks, and novel methods to enhance domain generalization and user experience. Moving forward, we envision agentic systems becoming central to next-generation recommender systems, offering seamless, trustworthy, and context-aware solutions across diverse applications. By tackling these challenges, researchers can push the boundaries of intelligent agent design and create systems that are not only innovative but also aligned with real-world needs and ethical considerations.

\bibliographystyle{ACM-Reference-Format}
\bibliography{sample-base}

\appendix

\end{document}